# Deep learning-based automated damage detection in concrete structures using images from earthquake events

Abdullah Türer[1,2], Yongsheng Bai[1,3], *Halil Sezen[1] and Alper Yilmaz[1]

[1] Department of Civil, Environmental and Geodetic Engineering, The Ohio State University, 2070 Neil Ave, Columbus, 43210, OH, USA
[2] Department of Civil Engineering, Ankara Yıldırım Beyazıt University, Takdir Street, 06010, Ankara, Türkiye
[3] Ingram School of Engineering, Texas State University, 327 W Woods Street, San Marcos, 78666, TX, USA

[1] sezen.1@osu.edu

## ABSTRACT

Timely assessment of integrity of structures after seismic events is crucial for public safety and emergency response. This study focuses on assessing the structural damage conditions using deep learning methods to detect exposed steel reinforcement in concrete buildings and bridges after large earthquakes. Steel bars are typically exposed after concrete spalling or large flexural or shear cracks. The amount and distribution of exposed steel reinforcement is an indication of structural damage and degradation. To automatically detect exposed steel bars, new datasets of images collected after the 2023 Turkey Earthquakes were labeled to represent a wide variety of damaged concrete structures. The proposed method builds upon a deep learning framework, enhanced with fine-tuning, data augmentation, and testing on public datasets. An automated classification framework is developed that can be used to identify inside/outside buildings and structural components. Then, a YOLOv11 (You Only Look Once) model is trained to detect cracking and spalling damage and exposed bars. Another YOLO model is fine-tuned to distinguish different categories of structural damage levels. All these trained models are used to create a hybrid framework to automatically and reliably determine the damage levels from input images. This research demonstrates that rapid and automated damage detection following disasters is achievable across diverse damage contexts by utilizing image data collection, annotation, and deep learning approaches.

## 1. INTRODUCTION

Large earthquakes can inflict significant damage on buildings and bridges, ranging from minor issues to partial or complete collapse. Visual inspection is vital for immediate safety assessments and forms the basis for informed decisions on building interventions, like repair, demolition, or reconstruction. It also plays a key role in accurately determining financial responsibilities among insurers, government agencies, or individual owners.

Deep learning techniques have demonstrated significant promise in recent years for facilitating quick and scalable damage detection. This advancement is built upon foundational work, including the development of extensive public image datasets for structural damage (Gao and Mosalam, 2018; Yeum et al., 2018) and the effective use of deep convolutional neural networks (CNNs) for image-level damage classification (Cha et al., 2018; Fan, 2024). More sophisticated studies have used various YOLO versions and object detection models like Faster R-CNN to localize multiple damage types simultaneously, including exposed rebar, cracks, and spalling (Bai et al., 2021a; Zou et al., 2022; Ghosh Mondal et al., 2020). Despite these developments, significant obstacles remain (Bai, 2022). First, even human inspectors find it difficult to discern damage levels, particularly when there are faint or obscure visual indicators. Second, diverse, well-annotated datasets representing a range of structural configurations, materials, lighting conditions, and damage types are necessary for reliable real-world performance. Third, few studies examine the structural significance of multiple damage types simultaneously, whereas many examine isolated damage, such as cracks or spalling.

This study aims to close these gaps by presenting a strong deep learning framework that automatically recognizes and categorizes earthquake-induced damage levels in reinforced concrete (RC) structures, with a particular emphasis on crucial damage indicators like cracking, spalling, and exposed rebar. The model used in this paper was trained using labeled image data from the 2023 Kahramanmaraş Earthquake in Türkiye and several benchmark datasets, including PEER Hub ImageNet ($\phi$-Net created by Gao and Mosalam (2018, 2020)) and publicly available crack-spalling datasets (Bai et al., 2021a, b). The generalization performance of the model was assessed using separate post-earthquake datasets from the 2017 Mexico Earthquake (Purdue University, 2018) and the 2017 Pohang Earthquake in South Korea (Sim et al., 2018). This data fusion allows us to train and test our models on diverse real-world scenarios and to improve the generalization capacity

## 2. METHODOLOGY

*2.1 Data preparation*

Four separate image datasets were used to train and evaluate the proposed framework, as summarized in Table 1. These datasets were developed using both publicly available resources (e.g., $\phi$-Net) and newly collected images from the 2023 Türkiye Earthquake.

Table 1 Summary of the image datasets used for training and evaluation.

| Dataset Type | Image Count | Resolution Range | Purpose | Notes |
| --- | --- | --- | --- | --- |

| Inside/Outside Classification | 18,193 | 224×224 – 1080×1440 | Classify indoor vs. outdoor | Based on ϕ-Net + Türkiye EQ |
|---|---|---|---|---|
| Structural Component Recognition | 4,939 | 224×224 – 1080×1240 | Identify beams, columns, and walls | Subset of ϕ-Net |
| Structural Damage Level Detection | 8,731 | 1080×1440 | Classify damage level (0–3) | Collected from Türkiye EQ |
| Damage Type Detection | 3,064 | Varying | Detect cracks, spalling, rebar | Bounding boxes annotated manually |

*2.2 Deep learning models for structural damage classification and detection*

YOLOv11, one of the latest iterations in the YOLO series by Ultralytics (Jocher and Qiu, 2024), marks a substantial leap forward in real-time object detection. The YOLO model allows for its application across various tasks: object detection, instance segmentation, image classification, pose estimation, and object tracking (Jegham et al., 2024).

*2.3 Hybrid deep learning framework for post-earthquake damage detection*

The framework is made up of a fusion logic after cascading YOLOv11-based classifiers and detectors. Prior to detecting structural elements and damage types (crack, spalling, and rebar), an image must first be classified as either inside or outside of a building. Lastly, a hybrid decision mechanism assigns one of four damage levels by combining predictions based on models and rules. Rebar validation and environment-aware filters are integrated into the updated RuleFusion v2, which improves accuracy without increasing the model size. Fig. 1 shows the overall workflow.

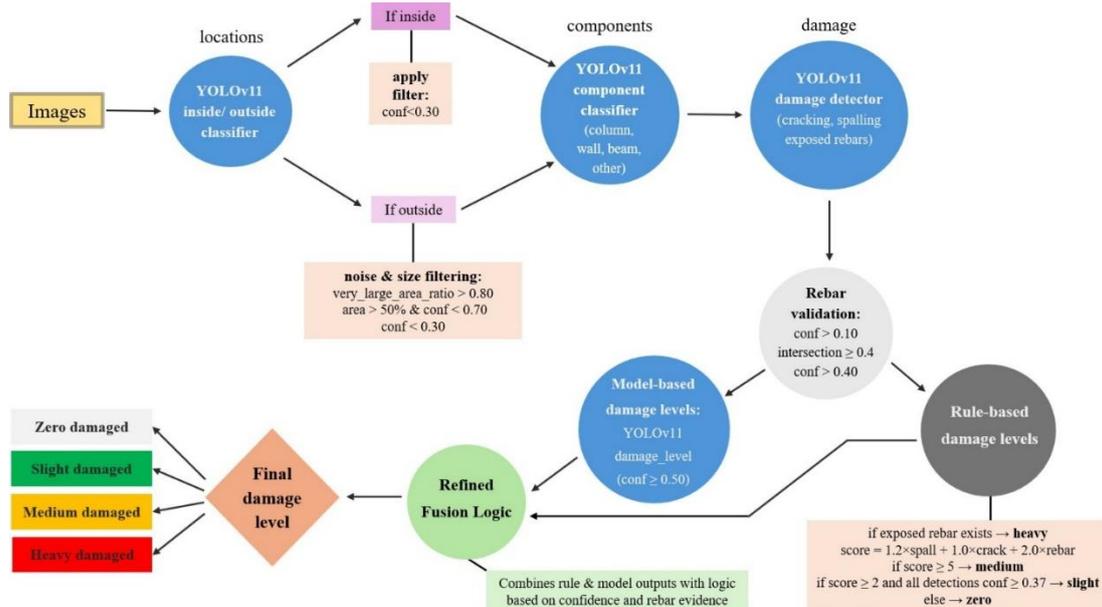

Fig. 1 Workflow of the proposed hybrid framework for post-earthquake building damage assessment

*2.4 Rule-based damage levels*

Initially, a rule-based classifier was created to improve reliability and interpretability. Damage is immediately classified as "heavy" if exposed rebar is found. If not, a weighted score is determined by counting the number of rebar, spalling, and cracks found as shown in Fig. 1. The damage is then categorized as zero, slight, or medium using thresholds. This reasoning guarantees consistent and comprehensible results, particularly under difficult visual circumstances.

## 3. RESULTS

The proposed hybrid framework was tested on 2017 Pohang Earthquake image (PEI) dataset (Sim et al., 2018) and 2017 Mexico City Earthquake image (MEI) dataset (Purdue-University, 2018), which include 4,109 and 4,136 high-resolution images collected by experts after these two Richter magnitude 5.2 and 7.1 earthquakes struck these regions. The model's overall performance was strong, but because of image quality, structural variations, lighting, and more complex context, its quantitative performance on the Mexico dataset was somewhat worse. Table 2 presents the accuracy results for different fusion and meta-model configurations. On the more difficult MEI dataset, however, the meta-model based on logistic regression performed poorly (Test No. 4). Consequently, an advanced LightGBM-based (Ke et al., 2017) meta-model was introduced, yet it only achieved an accuracy of 41.96% (±1 accuracy: 79.29%), indicating dataset-specific difficulties. These numerical insights suggest opportunities for further improvement through targeted hyperparameter tuning, feature engineering, or enriched training data in future studies.

Table 2 Accuracy comparison of the baseline hybrid model and the meta-model on different test datasets.

| Test | Dataset | Method | Model type | Accuracy (%) | ±1 Accuracy |
|---|---|---|---|---|---|
| 1st | PEI | Final Decision (Baseline) | Rule Fusion v1 | 61.54 | 68.32 |
| 2nd | PEI | Final Decision | Rule Fusion v2 | 71.04 | 91.92 |
| 3rd | PEI | Meta-Model Decision | Logistic Regression | 73.72 | 92.80 |
| 4th | MEI | Final Decision | Rule Fusion v2+LightGBM | 41.96 | 79.29 |

*\* v2 = v1 + environment-aware noise filtering, ambiguity-aware component bias, and refined rebar validation (see Section 2.3)*

The accuracy of the baseline fusion method on the 2017 PEI dataset was enhanced by incorporating a meta-model based on logistic regression, as indicated in Table 2. However, due to the previously mentioned dataset-specific complexities, neither Logistic

Regression nor an advanced LightGBM meta-model significantly improved performance on the more difficult 2017 MEI dataset.

Table 3 Per-class F1 scores on 2017 PEI dataset

| Metrics | Zero | Slight | Medium | Heavy |
|---|---|---|---|---|
| F1 Score | 0.844 | 0.384 | 0.128 | 0.641 |

Table 3 displays the per-class F1 scores of the top-performing configuration (Meta-Model Decision with Logistic Regression), which achieved 73.72% exact and 92.80% ±1 accuracy on the 2017 PEI dataset. The model is more accurate for zero and heavy damage classes, but less accurate for medium and slight damage levels.

## 4. CONCLUSIONS

This study presents a strong hybrid framework that combines deep learning, rule-based logic, and meta-learning for the precise classification of earthquake-induced structural damage. Key findings and contributions are summarized below:

(1) The suggested framework facilitates quick and accurate post-disaster evaluations by combining object detectors, image classifiers, and a fusion-based decision mechanism.

(2) The method successfully addressed class imbalance and produced results that were easy to understand by achieving high accuracy in identifying zero and heavy damage levels.

(3) The damage detection model demonstrates limitations in accurately identifying finer cracks, despite its success in detecting noticeable damage such as spalling and exposed rebar. This is because such damage looks obscured or subtle and is hard to discern from background textures.

Future research will concentrate on addressing performance limitations on some datasets by examining image-aware fusion techniques. The LightGBM or other machine learning algorithms will be incorporated into the scene-level embedding for the unmanned platforms with cameras.

**ACKNOWLEDGEMENTS**

This material is based upon work partially supported by the U.S. National Science Foundation under Grant No. 2036193. The first author acknowledges the support provided by Scientific and Technological Research Council of Türkiye (TÜBİTAK project number 1059B192401022) for his visit to the Ohio State University.

**REFERENCES**

Bai Y. (2022) Deep learning with vision-based technologies for structural damage detection and health monitoring. PhD Dissertation, The Ohio State University

Bai Y., Sezen H., and Yilmaz A. (2021a) Detecting cracks and spalling automatically in extreme events by end-to-end deep learning frameworks. ISPRS Annals of the Photogrammetry, Remote Sensing and Spatial Information Sciences 2:161–168

Bai Y., Sezen H., and Yilmaz A. (2021b) End-to-end deep learning methods for automated damage detection in extreme events at various scales. 25th International Conference on Pattern Recognition (ICPR) pp 6640–6647

Cha Y. J., Choi W., Suh G., Mahmoudkhani S., and Buyukozturk O. (2018) Autonomous structural visual inspection using region-based deep learning for detecting multiple damage types. Computer-Aided Civil and Infrastructure Engineering 33(9):731–747

Fan C. L. (2024) Deep neural networks for automated damage classification in imagebased visual data of reinforced concrete structures. Heliyon 10(19)

Gao Y., and Mosalam K. M. (2018) Deep transfer learning for image-based structural damage recognition. Computer-Aided Civil and Infrastructure Engineering 33(9):748–768

Gao Y., and Mosalam K. M. (2020) Peer hub imagenet: A large-scale multi-attribute benchmark data set of structural images. Journal of Structural Engineering 146(10):04020198

Ghosh Mondal T., Jahanshahi M. R., Wu R. T., and Wu Z. Y. (2020) Deep learning-based multi-class damage detection for autonomous post-disaster reconnaissance. Structural Control and Health Monitoring 27(4): e2507

Hosmer Jr D. W., Lemeshow S., and Sturdivant R. X. (2013) Applied logistic regression. John Wiley & Sons

Jegham N., Koh C. Y., Abdelatti M., and Hendawi A. (2024) Evaluating the evolution of yolo (you only look once) models: A comprehensive benchmark study of yolo11 and its predecessors. arXiv preprint arXiv:241100201

Jocher G., and Qiu J. (2024) Ultralytics yolo11. URL https://github.com/ultralytics/ultralytics

Ke G., Meng Q., Finley T., Wang T., Chen W., Ma W., Ye Q., and Liu T. Y. (2017) LightGBM: A highly efficient gradient boosting decision tree. In Advances in Neural Information Processing Systems 30

Purdue-University (2018) Buildings surveyed after 2017 Mexico City earthquakes. URL https://datacenterhub.org/resources/14746

Sim C., Laughery L., Chiou T. C., and Weng P. w. (2018) 2017 Pohang earthquake - reinforced concrete building damage survey. URL https://datacenterhub.org/resources/14728

Yeum C. M., Dyke S. J., and Ramirez J. (2018) Visual data classification in post-event building reconnaissance. Engineering Structures 155:16–24

Zou D., Zhang M., Bai Z., Liu T., Zhou A., Wang X., Cui W., and Zhang S. (2022) Multicategory damage detection and safety assessment of post-earthquake reinforced concrete structures using deep learning. Computer-Aided Civil and Infrastructure Engineering 37(9):1188–1204